%% file: arxiv_privilegednoise.tex
\documentclass[10pt]{article}
\usepackage{float}
\usepackage{rotating}
\usepackage{changepage}
\usepackage{nicefrac}
\usepackage[font=normalsize,format=plain,labelfont=bf,up,textfont=normal,up,justification=justified,singlelinecheck=false]{caption}
\include{definitions}

\usepackage{multirow}
\newcommand{\twocomine}[1]{\multicolumn{2}{c}{#1}}
\usepackage{verbatim}
\usepackage[toc,page]{appendix}
\usepackage{url}
\begin{document}
\title{Mind the Nuisance: Gaussian Process Classification\\ using Privileged Noise}

\author{Daniel~Hern\'{a}ndez-Lobato$^1$,~Viktoriia~Sharmanska$^2$,~Kristian~Kersting$^3$, \\Christoph~Lampert$^2$,
	and~Novi~Quadrianto$^4$\thanks{Corresponding Author: N.Quadrianto@sussex.ac.uk}	\vspace{0.5cm}\\
$^1$Universidad Aut\'{o}noma de Madrid, Spain\\
$^2$Institute of Science and Technology Austria, Austria\\
$^3$Technische Universit\"{a}t Dortmund, Germany\\
$^4$SMiLe CLiNiC, University of Sussex, UK}

\date{}

\maketitle

\begin{abstract}
The \emph{learning with privileged information} setting
has recently attracted a lot of attention within the machine
learning community, as it allows the integration of additional 
knowledge into the training process of a classifier, even when
this comes in the form of a data modality that is not available
at test time. Here, we show 
that privileged information can naturally be treated as noise in the 
latent function of a Gaussian Process classifier (GPC). That is, in contrast to 
the standard GPC setting, the latent function is not just a nuisance but
a feature: it becomes a natural measure of confidence about the training
data by modulating the slope of the GPC sigmoid likelihood function.
Extensive experiments on public datasets show that the proposed GPC method using privileged noise, 
called  GPC+, 
improves over a standard GPC without privileged knowledge, 
and also over the current state-of-the-art SVM-based method, SVM+. Moreover, we
show that advanced neural networks and deep learning methods 
can be compressed as privileged information.
\end{abstract}

\section{Introduction}
Prior knowledge is a crucial component of any learning system, 
as without a form of prior knowledge, learning is provably
impossible~\cite{wolpert1996lack}. 
Many forms of integrating prior knowledge into machine learning
algorithms have been developed: as a preference of certain prediction 
functions over others, as a Bayesian prior over parameters, or as 
additional information about the samples in the training set used 
for learning a prediction function.
In this work, we rely on the last of these setups, adopting Vapnik and Vashist's 
\emph{learning using privileged information (LUPI)}, see e.g.~\cite{VapVas09,PecVap10}:
{\it we want to learn a prediction function, \eg a classifier, and in 
addition to the main data modality that is to be used for prediction, 
the learning system has access to additional information about 
each training example.}

This scenario has recently attracted considerable interest within the
machine learning community, because it reflects well the increasingly 
relevant situation of \emph{learning as a service}: an expert 
trains a machine learning system for a specific task on request 
from a customer.
Clearly, in order to achieve the best result, the expert will use all 
the information available to him or her, not necessarily just the 
information that the system itself will have access to during its 
operation after deployment.
Typical scenarios for learning as a service include visual inspection tasks, in which a classifier 
makes real-time decisions based on the input from its sensor, but at 
training time, additional sensors could be made use of, and the processing 
time per training example plays less of a role. 
Similarly, a classifier built into a robot or mobile device operates 
under strong energy constraints, while at training time, 
energy is less of a problem, so additional data can be generated and 
made use of. 
A third, and increasingly important scenario is when the additional 
data is \emph{confidential}, as, \eg, in health care applications.
One can expect that a diagnosis system can be improved when more 
information is available at training time. One might, \eg, perform 
specific blood test, genetic sequence, or drug trials, for the 
subjects that form the training set.
However, the same data will not be available at test time, as 
obtaining it would be impractical, unethical, or outright illegal. 

In this work, we propose a novel method for using privileged information
based on the framework of Gaussian process classifiers (GPCs). 
The privileged data enters the model in form of a latent variable, 
which modulates the noise term of the GPC. 
Because the noise is integrated out before obtaining the final 
predictive model, the privileged information is indeed only 
required at training time, not at prediction time. 
The most interesting aspect of the proposed model is that 
by this procedure, the influence of the privileged information
becomes very interpretable: its role is to model the confidence 
that the Gaussian process has about any training example, 
which can be directly read off from the slope of the 
sigmoid-shaped GPC likelihood. 
Training examples that are \emph{easy} to classify by means of
their privileged data cause a faster increasing sigmoid, which 
means the GP trusts the taining example and tried to fit it well. 
Examples that are \emph{hard} to classify result in a slowly 
increase slope, so the GPC considers the training example less
reliable and puts not a lot of effort in fitting its label well.
Our experiments on multiple datasets show that this procedure 
leads not just to interpretable models, but also to significantly
higher classification accuracy. 

{\bf Related Work}
The LUPI framework was originally proposed by Vapnik and Vashist~\cite{VapVas09}, 
inspired by a thought-experiment: \emph{when training a soft-margin 
SVM, what if an \emph{oracle} would provide us with the optimal values 
of the slack variables?}
As it turns out, this would actually provably reduce the amount of training 
data needed, and consequently, Vapnik and Vashist proposed the \emph{SVM+} 
classifier that uses privileged data to predict values for the 
slack variables, which led to improved performance on several 
categorization tasks and found applications, \eg, in 
finance~\cite{ribeiro2010financial}.
This setup was subsequently improved, by a faster training 
algorithm~\cite{PecVap11}, better theoretical 
characterization~\cite{PecVap10}, and it 
was generalized, \eg, to the \emph{learning to rank} setting~\cite{ShaQuaLam13},
\emph{clustering}~\cite{feyereisl2012privileged}, 
and \emph{metric learning}~\cite{fouad2013incorporating}.
Recently, however, it was shown that the main effect of the SVM+ 
procedure is to assign a data-dependent weight to each training 
example in the SVM objective~\cite{LapHeiSch14}.
In contrast, GPC+ constitutes the first Bayesian treatment of classification using 
privileged information. Indeed, the resulting
\emph{privileged noise} approach is related to input-modulated noise commonly done in the regression task,
and several  Bayesian treatments of this \emph{heteroscedastic regression} using Gaussian processes
have been proposed. Since the predictive density and marginal likelihood are no longer
analytically tractable, most works in heteroscedastic GPs deal with approximate inference; techniques such as 
Markov Chain Monte Carlo \cite{GolWilBis98}, maximum a posteriori \cite{QuaKerReiCae09}, and 
recently a variational Bayes method \cite{GreTit11}. To our knowledge, however, there is no 
prior work on \emph{heteroscedastic classification} using GPs --- we will elaborate the reasons in Section \ref{sec:gpc} ---
and consequently this work develops the first approximate inference 
based on expectation propagation for the heteroscedastic noise case in the context of classification.

\section{GPC+: Gaussian Process Classification with Privileged Noise}
For self-consistency of the paper, we first review the GPC model \cite{RasWil06} with a particular 
emphasis on the noise-corrupted latent Gaussian process view. 
Then, we show how to treat privileged information as heteroscedastic noise in this latent process.
The elegant aspect of this view is the intuition as how the privileged noise is able to distinguish between \emph{easy} and \emph{hard} samples 
and in turn to re-calibrate our uncertainty in the original space. 

\subsection{Gaussian process classifier with noisy latent process}
\label{sec:gpc}
We are given a set of $N$ input-output data points or samples
$\Dcal = \{(\x_1,y_1),\ldots,(\x_N,y_N) \}\subset\RR^{d}\times\{0,1\}$.
Furthermore, we assume that the class label $y_i$ of sample $\x_i$ has been generated as 
$y_n = \I[\ \tildef(\x_n)\geq 0\ ]$, where $\tildef(\cdot)$ is a \emph{noisy} latent function and $\I[\cdot]$ is the Iverson's bracket 
notation: $\I[\ P\ ]=1$ when the condition $P$ is true, and $0$ otherwise.
Induced by the label generation process, we adopt the following form of likelihood function 
for $\bftildef = (\tildef(\x_1),\ldots,\tildef(\x_N))^\top$:
\begin{align}
\pr(\y|\bftildef,X=(\x_1,\ldots,\x_N)^\top) = \prod\nolimits_{n=1}^{N} \pr(y_n=1|\x_n,\tildef) = \prod\nolimits_{n=1}^N \I[\ \tildef(\x_n)\geq 0\ ],
\end{align}  
where the noisy latent function at sample $\x_n$ is given by $\tildef(\x_n)= f(\x_n) + \epsilon_n$ 
with $f(\x_n)$ being the \emph{noise-free} latent function. The noise term $\epsilon_n$ is 
assumed to be independent and normally distributed with zero mean and variance $\sigma^2$, that is $\epsilon_n\sim\Ncal(\epsilon_n|0,\sigma^2)$. 
To make inference about $\tildef(\x_n)$, we need to specify a prior over this function. 
We proceed by imposing a zero mean Gaussian process prior \cite{RasWil06} on the noise-free latent function, that is $f(\x_n) \sim \mathcal{GP}(0,k(\x_n,\cdot))$ 
where $k(\cdot,\cdot)$ is a positive-definite kernel function \cite{SchSmo01} that specifies prior properties of $f(\cdot)$.
A typical kernel function that allows for non-linear smooth function is the squared exponential kernel 
$k_f(\x_n,\x_m) = \theta\exp(-\frac{1}{2l}\nbr{\x_n-\x_m}^2_{\ell_2})$. In this kernel function, the parameter $\theta$ controls 
the amplitude of function $f(\cdot)$ while $l$ controls the smoothness of $f(\cdot)$. 
Given the prior and the likelihood, Bayes' rule is used to compute the posterior of $\tildef(\cdot)$, 
that is $\pr(\bftildef|X,\y) = \nicefrac{\pr(\y|\bftildef,X)\pr(\bftildef)}{\pr(\y|X)}$.

We can simplify the above noisy latent process view by integrating out the noise term $\epsilon_n$ and writing down the 
individual likelihood at sample $\x_n$ in term of noise-free latent function $f(\cdot)$ as follows 
\begin{align}
\label{eq:probitlikelihood}
\pr(y_n=1|\x_n,f) &= \!\!\int\!\! \I[\tildef(\x_n)\geq 0] \Ncal(\epsilon_n|0,\sigma^2) d\epsilon_n = \!\!\int \!\!\I[\epsilon_n \geq -f(\x_n)] \Ncal(\epsilon_n|0,\sigma^2) d\epsilon_n\nonumber\\ &= \Phi_{(0,\sigma^2)}(f(\x_n)),
\end{align}
where $\Phi_{(\mu,\sigma^2)}(\cdot)$ is a Gaussian cumulative distribution function (CDF) with mean $\mu$ and variance $\sigma^2$. 
Typically the standard Gaussian CDF is used, that is $\Phi_{(0,1)}(\cdot)$, in the likelihood of \eq{eq:probitlikelihood}. 
Coupled with a Gaussian process prior on the latent function $f(\cdot)$, this results in the widely adopted noise-free latent Gaussian process view with probit likelihood. 
The equivalence between a noise-free latent process with probit likelihood and a noisy latent process with step-function likelihood is widely known \cite{RasWil06}.
It is also widely accepted that the noisy latent function $\tildef$ (or the noise-free latent function $f$) is a \emph{nuisance} function as we do not observe the value of this function itself 
and its sole purpose is for a convenient formulation of the classification model \cite{RasWil06}.
However, in this paper, we show that by using privileged information as the noise term, the latent function $\tildef$ now plays a crucial role.
The latent function with privileged noise adjusts the slope transition in the Gaussian CDF to be \emph{faster} or \emph{slower} corresponding to more \emph{certainty} or more \emph{un}certainty 
about the samples in the original space. This is described in details in the next section. 

\subsection{Privileged information is in the Nuisance Function}
In the learning using privileged information (LUPI) paradigm \cite{VapVas09}, besides input data points $\{\x_1,\ldots,\x_N \}$ and 
associated outputs $\{y_1,\ldots,y_N\}$, we are given additional information $\x^*_n \in \RR^{d^*}$ about each training instance $\x_n$. 
However this privileged information will not be available for unseen test instances. 
Our goal is to exploit the additional data $\x^*$ to influence our choice of the latent function $\tildef(\cdot)$. 
This needs to be done while making sure that the function does not directly use the privileged data as input, as it is simply not available at test time. 
We achieve this naturally by treating the privileged information as a heteroscedastic (input-dependent) noise in the latent process. 

Our classification model with privileged noise is then as follows:
\begin{align}
\text{Likelihood model}: &\ \pr(y_n=1|\x_n,\tildef) = \I[\ \tildef(\x_n)\geq 0\ ]\quad\text{where}\quad\x_n\in\RR^d\\
\text{Assume}: &\ \tildef(\x_n)= f(\x_n) + \epsilon_n\\
\text{Privileged noise model}: &\ \epsilon_n\overset{i.i.d.}{\sim}\Ncal(\epsilon_n|0,z(\x^*_n) = \exp(g(\x^*_n)))\quad\text{where}\quad\x^*_n\in\RR^{d^*}\\
\text{GP prior model}: &\ f(\x_n) \sim \mathcal{GP}(0,k_f(\x_n,\cdot)) \quad \text{and}\quad g(\x^*_n) \sim \mathcal{GP}(0,k_g(\x^*_n,\cdot)).
\end{align}
In the above, the $\exp(\cdot)$ function is needed to ensure positivity of the noise variance. 
The term $k_g(\cdot,\cdot)$ is a positive-definite kernel function that specifies the prior properties of another latent function $g(\cdot)$ 
which is evaluated in the privileged space $\x^*$. 
Crucially, the noise term $\epsilon_n$ is now \emph{heteroscedastic}, that is it has a different variance $z(\x^*_n)$ at each input point $\x_n$.
This is in contrast to the standard GPC approach discussed in Section \ref{sec:gpc} where 
the noise term is assumed to be homoscedastic, $\epsilon_n\sim\Ncal(\epsilon_n|0,z(\x^*_n)=\sigma^2)$. 
Indeed, an input-dependent noise term is very common in a task with continuous output values $y_n\in\RR$ (a regression task), resulting in the so-called heteroscedastic regression models, which have been proven 
to be more flexible in numerous applications as already touched upon in the related work section. 
However, to our knowledge, there is no prior work on \emph{heteroscedastic classification} models. This is not surprising as the nuisance view of the latent function renders having a flexible input-dependent noise point-less.

\begin{figure*}[tbh]
\begin{center}\quad \ \
\begin{minipage}{0.45\linewidth}
    \includegraphics[width=1\columnwidth]{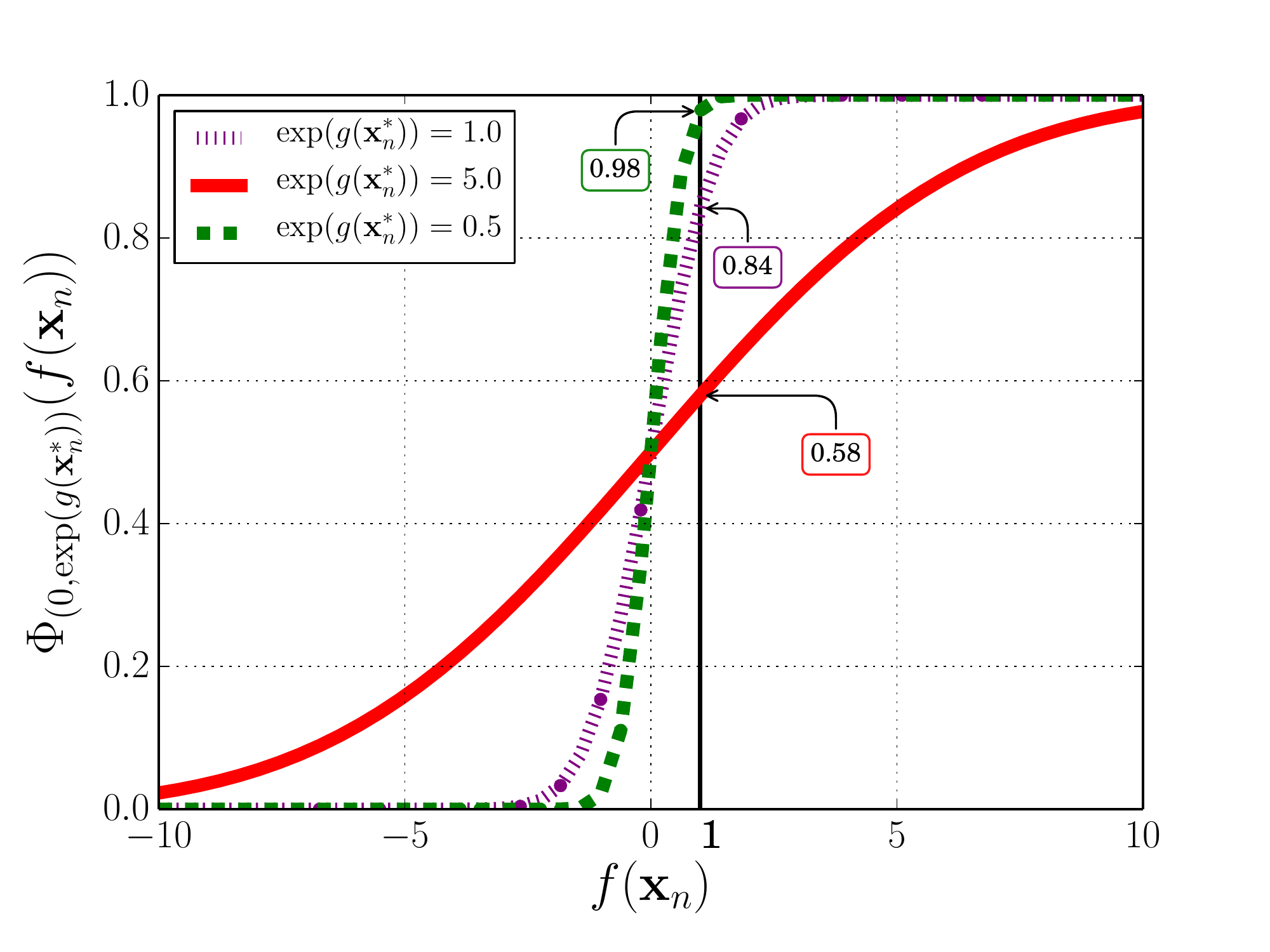}
\end{minipage}\hfill
\begin{minipage}{0.45\linewidth}
    \includegraphics[width=1\columnwidth]{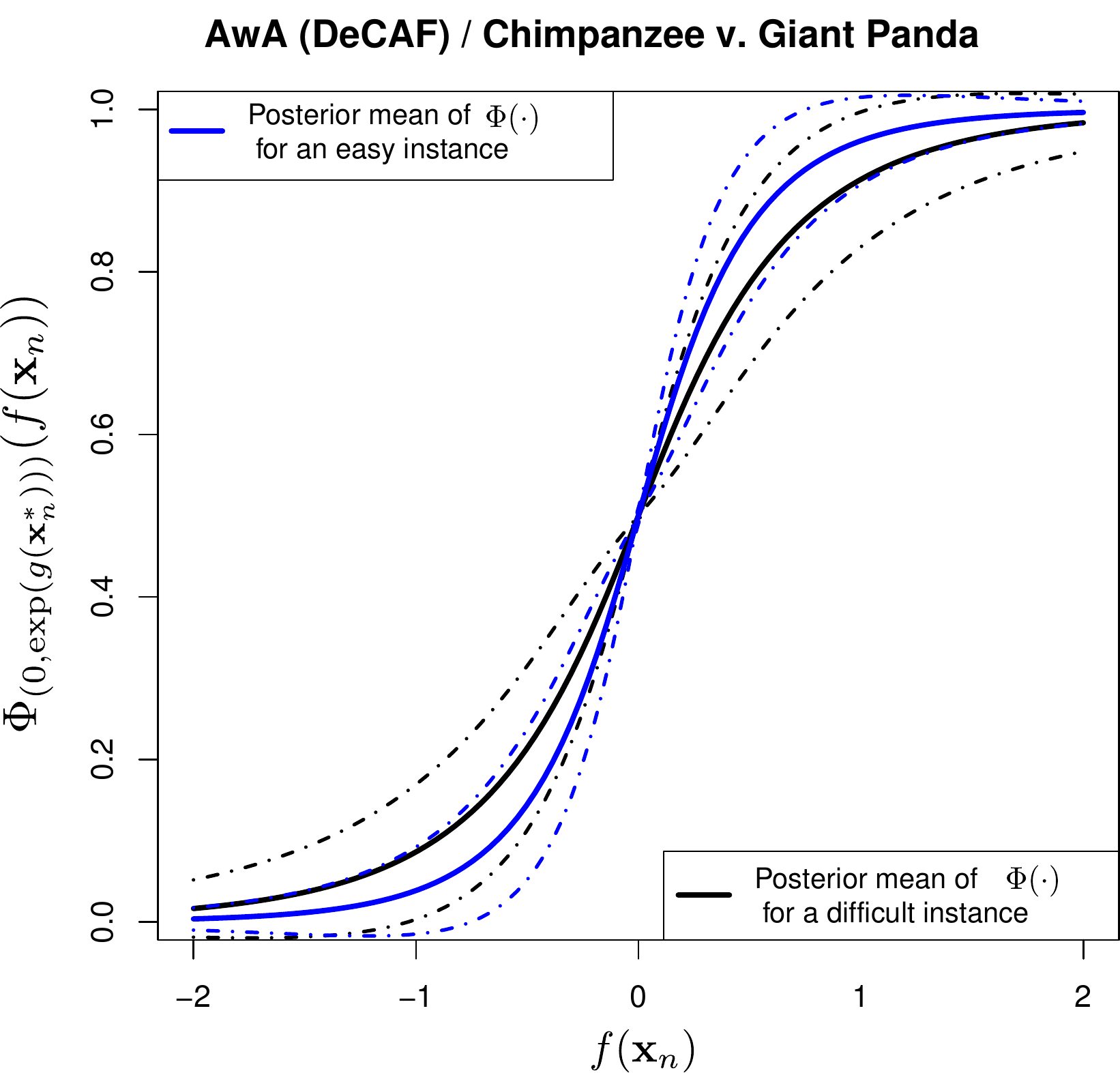}
\end{minipage}
  \caption{Effects of privileged noise on the nuisance function. {\bf (Left)}  On synthetic data. Suppose for an input $\x_n$, the latent function value is $f(\x_n) = 1$. Now also assume that the associated privileged information $\x_n^*$ for $n$-th data point deems the sample as \emph{difficult}, say $\exp(g(\x^*_n)) = 5.0$. Then 
  the likelihood will reflect this uncertainty $\pr(y_n=1|f,g,\x_n,\x_n^*) = 0.58$. In contrast, if the associated privileged information considers the sample as \emph{easy}, say e.g. $\exp(g(\x^*_n)) = 0.5$, the likelihood is very certain $\pr(y_n=1|f,g,\x_n,\x_n^*) = 0.98$. {\bf (Right)} On real data taken from our experiments in Sec.~\ref{sec:experiments}. The posterior means of the 
  $\Phi(\cdot)$ function (solid) and its $1$-standard deviation confidence interval (dash-dot) for easy (blue) and difficult (black) instances of the Chimpanzee v. Giant Panda binary task on the Animals with Attributes (AwA) dataset. (Best viewed in color)\label{fig:gaussiancdf}}
\end{center}
\end{figure*}
In the context of learning with privileged information, however, heteroscedastic classification 
is actually a very sensible idea.  
This is best illustrated when investigating 
the effect of privileged information in the equivalent formulation of a noise free latent process, i.e., one integrates out the privileged input-dependent noise term:
\begin{align}
\pr(y_n=1|\x_n,\x^*_n,f,g) &= \!\!\int \I[\ \tildef(\x_n)\geq 0\ ] \Ncal(\epsilon_n|0,\exp(g(\x^*_n)) d\epsilon_n \\
			 &= \!\!\int \I[\ \epsilon_n \leq f(\x_n)\ ] \Ncal(\epsilon_n|0,\exp(g(\x^*_n)) d\epsilon_n \\
                  	 &= \Phi_{(0,\exp(g(\x^*_n)))}(f(\x_n)) = \Phi_{(0,1)}(f(\x_n)/\sqrt{\exp(g(\x^*_n)})\label{eq:heteroprobitlikelihood}.
\end{align}
From \eq{eq:heteroprobitlikelihood}, it is clear that privileged information adjusts the slope transition of the Gaussian CDF. For difficult (a.k.a. noisy) samples, the latent function $g(\cdot)$ will be high, the slope transition will be slower, 
and thus more uncertainty is in the likelihood term $\pr(y_n=1|\x_n,\x^*_n,f,g)$. For easy samples, however, the latent function $g(\cdot)$ will be low, the slope transition will be faster, and thus less uncertainty is in the likelihood term $\pr(y_n=1|\x_n,\x^*_n,f,g)$. This is illustrated in Fig.~\ref{fig:gaussiancdf}.

\subsection{Posterior and Prediction on Test Data}
Define $\mathbf{g}=(g(\mathbf{x}_1^*),\ldots,g(\mathbf{x}_n^*))^\text{T}$
and $\mathbf{X}^*=(\mathbf{x}_1^*,\ldots,\mathbf{x}_n^*)^\text{T}$.
Given the conditional \emph{i.i.d.} likelihood $\pr(\mathbf{y}|\mathbf{X},\mathbf{X}^\star,\mathbf{f},\mathbf{g}) = \prod_{n=1}^{N} \pr(y_n=1|f,g,\x_n,\x_n^*)$ with the per observation likelihood term $\pr(y_n|f,g,\x_n,\x_n^*)$ given in \eq{eq:heteroprobitlikelihood} and the Gaussian process priors on functions, the posterior for $\mathbf{f}$ and $\mathbf{g}$ is:
\begin{align}
\pr(\mathbf{f},\mathbf{g}|\mathbf{y},\mathbf{X},\mathbf{X}^\star) & = 
	\frac{\pr(\mathbf{y}|\mathbf{X},\mathbf{X}^\star,\mathbf{f},\mathbf{g})\pr(\mathbf{f})\pr(\mathbf{g})}
	{\pr(\mathbf{y}|\mathbf{X},\mathbf{X}^\star)} \,,
\label{eq:posterior}
\end{align}
where $\pr(\mathbf{y}|\mathbf{X},\mathbf{X}^\star)$ can be maximised with respect to a set of hyper-parameter values such as amplitude $\theta$ and smoothness $l$ parameters of the kernel functions \cite{RasWil06}. 
For a previously unseen test point $\mathbf{x}_\text{new}\in\RR^d$, the predictive distribution for its label $y_\text{new}$ is given as:
\begin{align}
\pr(y_\text{new}=1|\mathbf{y},\mathbf{X},\mathbf{X}^\star) &= 
\int \I[\ \tildef(\mathbf{x}_\text{new})\geq 0\ ] \pr(f_\text{new}|\mathbf{f})
\pr(\mathbf{f},\mathbf{g}|\mathbf{y},\mathbf{X},\mathbf{X}^\star) d \mathbf{f} d\mathbf{g} d f_\text{new}\,,
\label{eq:predictive}
\end{align}
where $\pr(f_\text{new}|\mathbf{f})$ is a Gaussian conditional distribution. We note that in \eq{eq:predictive}, 
we do not consider the privileged information $\mathbf{x}^\star_\text{new}$ associated 
to $\mathbf{x}_\text{new}$. 
The interpretation is we consider a homoscedastic noise at test time. This is a reasonable approach as there is no additional information for increasing or decreasing our confidence in the newly observed data $\mathbf{x}_\text{new}$.
Finally, we predict the label for a test point via Bayesian decision theory: the label being predicted is the one 
with the largest probability. 

\section{Expectation Propagation with Numerical Quadrature}
Unfortunately, as for most interesting Bayesian models, inference in the GPC+ model is 
very challenging. Already in the homoscedastic case, the predictive density and marginal
likelihood are not  analytically tractable. 
In this work, we therefore adapt Minka's
expectation propagation (EP) \cite{Minka01} with numerical quadrature for approximate inference.
Please note that EP is
the preferred method for approximate inference with GPCs in terms of 
accuracy and computational cost \cite{NicRas08,kuss2005}. 

Consider the 
joint distribution of $\mathbf{f}$, $\mathbf{g}$ and $\mathbf{y}$. Namely,
$\text{Pr}(\mathbf{y}|\mathbf{X},\mathbf{X}^*,\mathbf{f},\mathbf{g})\text{Pr}(\mathbf{f})\text{Pr}(\mathbf{g})$
where $\text{Pr}(\mathbf{f})$ and $\text{Pr}(\mathbf{g})$ are 
Gaussian process priors and the likelihood $\text{Pr}(\mathbf{y}|\mathbf{X},\mathbf{X}^*,\mathbf{f},\mathbf{g})$ 
is equal to $\prod_{n=1}^N \text{Pr}(y_n|\mathbf{x}_n,\mathbf{x}_n^*,f,g)$, with $\text{Pr}(y_n|\mathbf{x}_n,\mathbf{x}_n^*,f,g)$ 
given by (\ref{eq:heteroprobitlikelihood}). EP approximates each non-normal factor in this joint distribution 
by an un-normalised bi-variate normal distribution of $f$ and $g$ (we assume independence between $f$ and $g$). 
The only non-normal factors correspond to those of the likelihood. These are approximated as:
\begin{align}
\text{Pr}(y_n|\mathbf{x}_n,\mathbf{x}_n^*,f,g) & \approx \overline{\gamma}_n(f,g) = \overline{z}_n 
	\mathcal{N}(f(\mathbf{x}_n)|\overline{m}_f,\overline{v}_f) \mathcal{N}(g(\mathbf{x}_n^*)|\overline{m}_g,\overline{v}_g)\,,
\end{align}
where the parameters with the super-script $\overline{\,^{\,}}$ are to be found by EP. The posterior approximation 
$\mathcal{Q}$ computed by EP results from normalising with respect to $\mathbf{f}$ and $\mathbf{g}$ the EP approximate joint
distribution. This distribution $\mathcal{Q}$ is obtained by replacing each likelihood factor by the corresponding approximate 
factor $\overline{\gamma}_n$. In particular,
\begin{align}
\text{Pr}(\mathbf{f},\mathbf{g}|\mathbf{X},\mathbf{X}^*,\mathbf{y}) \approx \mathcal{Q}(\mathbf{f},\mathbf{g})
:= Z^{-1}[\prod\nolimits_{n=1}^N\overline{\gamma}(f,g)]\text{Pr}(\mathbf{f})\text{Pr}(\mathbf{g})\,,
\end{align}
where $Z$ is a normalisation constant that approximates the model evidence $\text{Pr}(\mathbf{y}|\mathbf{X},\mathbf{X}^*)$. 
The normal distribution belongs to the exponential family of probability distributions and is closed under the 
product and division. It is hence possible to show that $\mathcal{Q}$ is the product of two multi-variate normals \cite{matthias2006}.
The first normal approximates the posterior for $\mathbf{f}$ and the second the posterior for $\mathbf{g}$.

EP tries to fix the parameters of $\overline{\gamma}_n$ so that it is similar to the exact factor
$\text{Pr}(y_n|\mathbf{x}_n,\mathbf{x}_n^*,f,g)$ in regions of high posterior probability \cite{Minka01}. 
For this, EP iteratively updates each $\overline{\gamma}_n$ until convergence to minimise 
$\text{KL}\left(\text{Pr}(y_n|\mathbf{x}_n,\mathbf{x}_n^\star,f,g)\mathcal{Q}^\text{old} / Z_n||\mathcal{Q}\right)$,
where $\mathcal{Q}^\text{old}$ is a normal distribution proportional to 
$\left[\prod_{n'\neq n} \overline{\gamma}_{n'}\right]\text{Pr}(\mathbf{f})\text{Pr}(\mathbf{g})$
with all variables different from $f(\mathbf{x}_n)$ and $g(\mathbf{x}_n^*)$ marginalised out,
$Z_n$ is simply a normalisation constant and $\text{KL}(\cdot||\cdot)$ denotes the Kullback-Leibler divergence
between probability distributions. Assume $\mathcal{Q}^\text{new}$ is the distribution minimising the previous divergence.
Then, $\overline{\gamma}_n \propto \mathcal{Q}^\text{new} / \mathcal{Q}^\text{old}$ 
and the parameter $\overline{z}_n$ of $\overline{\gamma}_n$ is fixed to guarantee that $\overline{\gamma}_n$ integrates
the same as the exact factor with respect to $\mathcal{Q}^\text{old}$.
The minimisation of the KL divergence involves matching expected sufficient statistics (mean and variance) between 
$\text{Pr}(y_n|\mathbf{x}_n,\mathbf{x}_n^\star,f,g)\mathcal{Q}^\text{old} / Z_n$ and $\mathcal{Q}^\text{new}$. 
These expectations can be obtained from the derivatives of $\log Z_n$ with respect to the (natural) parameters 
of $\mathcal{Q}^\text{old}$ \cite{matthias2006}. Unfortunately, the computation of $\log Z_n$ in closed form is intractable. 
We show here that it can be approximated by a \emph{one dimensional quadrature}. Denote by 
$m_f$, $v_f$, $m_g$ and $v_g$ the means and variances of $\mathcal{Q}^\text{old}$ for $f(\mathbf{x}_n)$ 
and $g(\mathbf{x}_n^*)$, respectively. Then,
\begin{align}
Z_n = \int \Phi_{(0,1)}\left(\frac{y_n m_f}{\sqrt{v_f + \exp(g(\mathbf{x}_n^*))}} \right)
\mathcal{N}(g(\mathbf{x}_n^*)|m_g,v_g) d g(\mathbf{x}_n^*)\,.
\end{align}
Thus, the EP algorithm only requires five quadratures to update each $\overline{\gamma}_n$. 
A first one to compute $\log Z_n$ and four extras to 
compute its derivatives with respect to $m_f$, $v_f$, $m_g$ and $v_g$. After convergence, $\mathcal{Q}$ can 
be used to approximate predictive distributions and the normalisation constant $Z$ can be maximised to find 
good values for the model's hyper-parameters. 
In particular, it is possible to compute the gradient of $Z$ with respect to the parameters of the Gaussian process
priors for $\mathbf{f}$ and $\mathbf{g}$ \cite{matthias2006}. 

\section{Experiments}
\label{sec:experiments}
Our intention here is to investigate the performance of the 
GP with privileged noise approach. To this aim, we considered three types of binary classification tasks 
corresponding to different privileged information using two real-world datasets: \emph{Attribute Discovery} and \emph{Animals with Attributes}. 
We detail those experiments in turn in the following sections.

{\bf Methods} We compared our proposed \texttt{GPC+} method with the 
well-established LUPI method based on SVM, \texttt{SVM+} \cite{PecVap11}.
As a reference, we also fit standard GP and SVM classifiers when learning on the original space $\RR^d$
(\texttt{GPC} and \texttt{SVM} baselines). For \emph{all four} methods, we used a squared exponential kernel with amplitude parameter 
$\theta$ and smoothness parameter $l$. For simplicity, we set $\theta = 1.0$ in all cases. For GPC and GPC+, 
we used type II-maximum likelihood for estimating the hyper-parameters. There are two 
hyper-parameters in GPC (smoothness parameter $l$ and noise variance $\sigma^2$) and also two in GPC+ 
(smoothness parameters $l$ of kernel $k_f(\cdot,\cdot)$ and of kernel $k_g(\cdot,\cdot)$).
For SVM and SVM+, we used cross-validation to set the hyper-parameters. SVM has 
two knobs, that is smoothness and regularisation, and SVM+ has four knobs, 
two smoothness and two regularisation parameters. 
It turned out that a grid search via cross validation was too expensive for searching the best parameters in SVM+, 
we instead use the performance on a separate validation set to guide the search process.
None of the other three methods used this separate validation set, this means that 
we give a competitive advantage to SVM+ over the other methods.

{\bf Evaluation metric}
To evaluate the performance of the methods we used classification error on an independent test set.
We performed $100$ repeats of all the experiments to get the better statistics of the performance and report the mean and the standard error.

\subsection{Attribute Discovery Dataset \cite{BerBerShi10}}
\label{sec:attributediscovery}
The data set was collected from a shopping website that aggregates product data from variety of e-commerce sources and includes both images and associated textual descriptions. The images and associated texts are grouped into $4$ broad shopping categories: \textit{bags, earrings, ties, and shoes}. 
We used $1800$ samples from this dataset.
We generated $6$ binary classification tasks for each pair of the $4$ classes with $200$ samples for training, $200$ samples for validation, 
and the rest of samples for testing the predictive performance.

{\bf Neural networks on texts as privileged information}
We used \emph{images} as the \emph{original} domain and \emph{texts} as the \emph{privileged} domain.
This setting was also explored in \cite{ShaQuaLam13}. However, we used a different dataset as textual descriptions of the images used in 
\cite{ShaQuaLam13} are sparse and contain duplicates. Furthermore, we extracted more advanced text features instead of simple term frequency (TF) features.
As image representation, we extracted SURF descriptors \cite{BayEssTuyVan08} 
and constructed a codebook of $100$ visual words using the $k$-means clustering.
As text representation, we extracted $200$ dimensional continuous word-vector representation using a neural network skip-gram architecture \cite{MikSutCheCoretal13}\footnote{\url{https://code.google.com/p/word2vec/}}.
To convert this word representation to a fixed-length sentence representation, we constructed a codebook of $100$ word-vector using again $k$-means clustering.
We note that a more elaborate approach to transform word to sentence or document features has recently been developed \cite{LeMik14}, and we are planning to explore this in the future. We performed PCA for dimensionality reduction in the original and privileged domains and only kept the top $50$ principal components. 
Finally, we standardised the data so that each feature has zero mean and unit standard deviation. 

The experimental results are summarised in Tab.~\ref{tab:ADNNtexts}. On average over $6$ tasks, SVM with hinge loss outperforms GPC with probit likelihood. 
However, GPC+ significantly improves over GPC providing the best results on average. This clearly shows that GPC+ is able to utilise the neural network textual representation as privileged information. In contrast,
SVM+ produced the same result as SVM. We suspect this is due to: SVM has already shown strong performance on the original image space coupled with the difficulties 
in finding the best values of four hyper-parameters. Keep in mind that, in SVM+, we discretised the hyper-parameter search space over $625$ ($5\times 5\times 5\times 5$) possible combination values and used a separate validation technique.

\begin{table}[tb]
\centering
\scalebox{1.0}{
\begin{tabular}{l*{5}{r@{$\pm$}l}}
      		             & \twocomine{\texttt{GPC}} & \twocomine{\texttt{GPC+ (Ours)}} & \twocomine{\texttt{SVM}} & \twocomine{\texttt{SVM+}}\\
\hline
\texttt{ bags v. earrings}   & 9.79&0.12    & {\bf 9.50}&{\bf 0.11}     & 9.89&0.14     & 9.89&0.13\\
\texttt{ bags v. ties}   	 & 10.36&0.16   & 10.03&0.15    & {\bf 9.44}&{\bf 0.16}     & 9.47&0.13\\
\texttt{  bags v. shoes}   	 & 9.66&0.13    & {\bf 9.22}&{\bf 0.11}     & 9.31&0.12     & 9.29&0.14\\
\texttt{ earrings v. ties}   & 10.84&0.14   & {\bf 10.56}&{\bf 0.13}    & 11.15&0.16    & 11.11&0.16\\
\texttt{ earrings v. shoes}  & 7.74&0.11    & {\bf 7.33}&{\bf 0.10}     & 7.75&0.13     & 7.63&0.13\\
\texttt{ ties v. shoes}   	 & 15.51&0.16   & 15.54&0.16    & {\bf 14.90}&{\bf 0.21}    & 15.10&0.18\\
\hline
{\it average error}      & 10.65&0.11       & {\bf10.36}&{\bf0.12}  & 10.41&0.11        & 10.42&0.11\\
{\it average ranking}            & \twocomine{3.0}  & \twocomine{\bf 1.8}       & \twocomine{2.7}   & \twocomine{2.5}
\end{tabular}}
\caption{Error rate performance (the lower the better) on the Attribute Discovery dataset over 100 repeated experiments. We used \textbf{images} as the \textbf{original} domain and neural networks word-vector representation on \textbf{texts} as the \textbf{privileged} domain. The best method for each binary task is highlighted in \textbf{boldface}. An average rank equal to one means that the corresponding method has the smallest error on the $6$ tasks.\label{tab:ADNNtexts}
}
\end{table}

\begin{figure}[tbh]
\begin{center}
\textbf{DeCAF as privileged}\\
  \includegraphics[width=0.8\columnwidth, page=1]{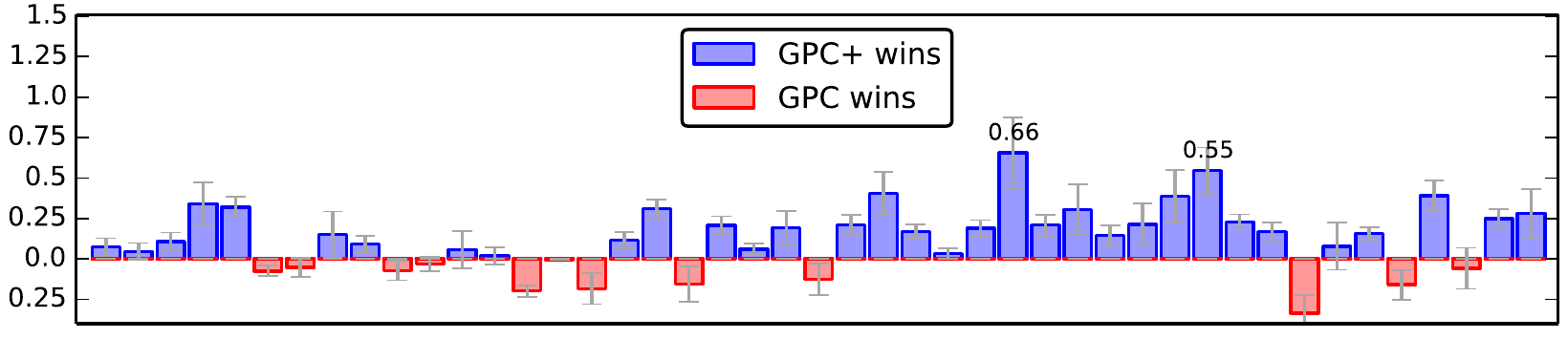}
  \includegraphics[width=0.8\columnwidth, page=2]{plotDECAF.pdf}\\
  \textbf{Attributes as privileged}\\
    \includegraphics[width=0.8\columnwidth, page=1]{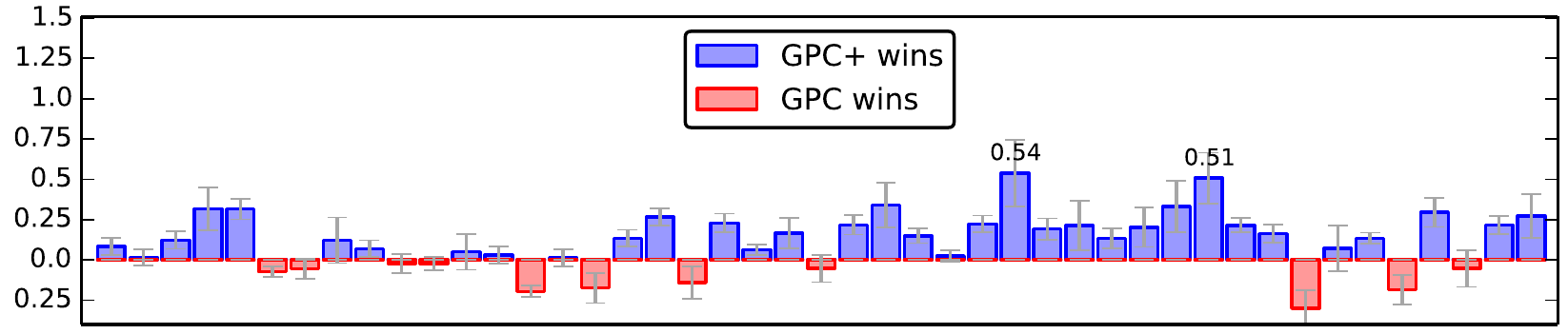}
    \includegraphics[width=0.8\columnwidth, page=2]{plotDAP.pdf}
  \caption{Pairwise comparison of the proposed GPC+ method and main baselines is shown via relative difference of the error rate (top: GPC+ versus GPC, bottom: GPC+ versus SVM+). The length of the $45$ bars corresponds to relative difference of the error rate over $45$ cases. Full results of the average error rate performance on AwA dataset across $45$ tasks over $100$ repeated experiments are in the appendix. (Best viewed in color)\label{fig:pairwise}} 
\end{center}
\end{figure}

\subsection{Animals with Attributes (AwA) Dataset \cite{LamNicHar14}}
The dataset was collected by querying the image search engines for each of the $50$ animals categories 
which have complimentary high level description of the semantic properties such as shape, colour, or habitation forms, among others.
The semantic attributes per animal class were retrieved from a prior psychological study. 
We focused on the $10$ categories corresponding to the test set of this dataset for 
which the predicted attributes are 
provided based on the probabilistic DAP model \cite{LamNicHar14}.
The $10$ classes are: \textit{chimpanzee, giant panda, leopard, persian cat, pig, hippopotamus, humpback whale, raccoon, rat, seal}, and contain $6180$ images in total. 
As in Section \ref{sec:attributediscovery} and also in \cite{ShaQuaLam13}, we generated $45$ binary classification tasks for each pair of the $10$ classes with $200$ samples for training, $200$ samples for validation, 
and the rest of samples for testing the predictive performance.

{\bf Neural networks on images as privileged information}
Deep learning methods have gained an increased attention within the machine learning and computer vision community over the recent years.
This is due to their capability in extracting informative features and delivering strong predictive performance in many classification tasks.
As such, we are interested to explore the use of deep learning based features as privileged information
so that their predictive power can be used even if we do not have access to them at prediction time. 
We used the standard \emph{SURF} features \cite{BayEssTuyVan08} with $2000$ visual words as the \emph{original} domain and 
used the recently proposed \emph{DeCAF} features \cite{DonJiaVinHofetal14} extracted from the activation of a deep convolutional network trained in a fully supervised fashion as the \emph{privileged} domain. The DeCAF features were in $4096$ dimensions. 
All features are provided with the AwA dataset\footnote{\url{http://attributes.kyb.tuebingen.mpg.de}}. 
We again performed PCA for dimensionality reduction in the original and privileged domains and only kept the top $50$ principal components, as well as standardised the data. 

\begin{figure*}[tp]
\begin{center}
\begin{tabular}{c|c}
    \includegraphics[width=0.45\columnwidth]{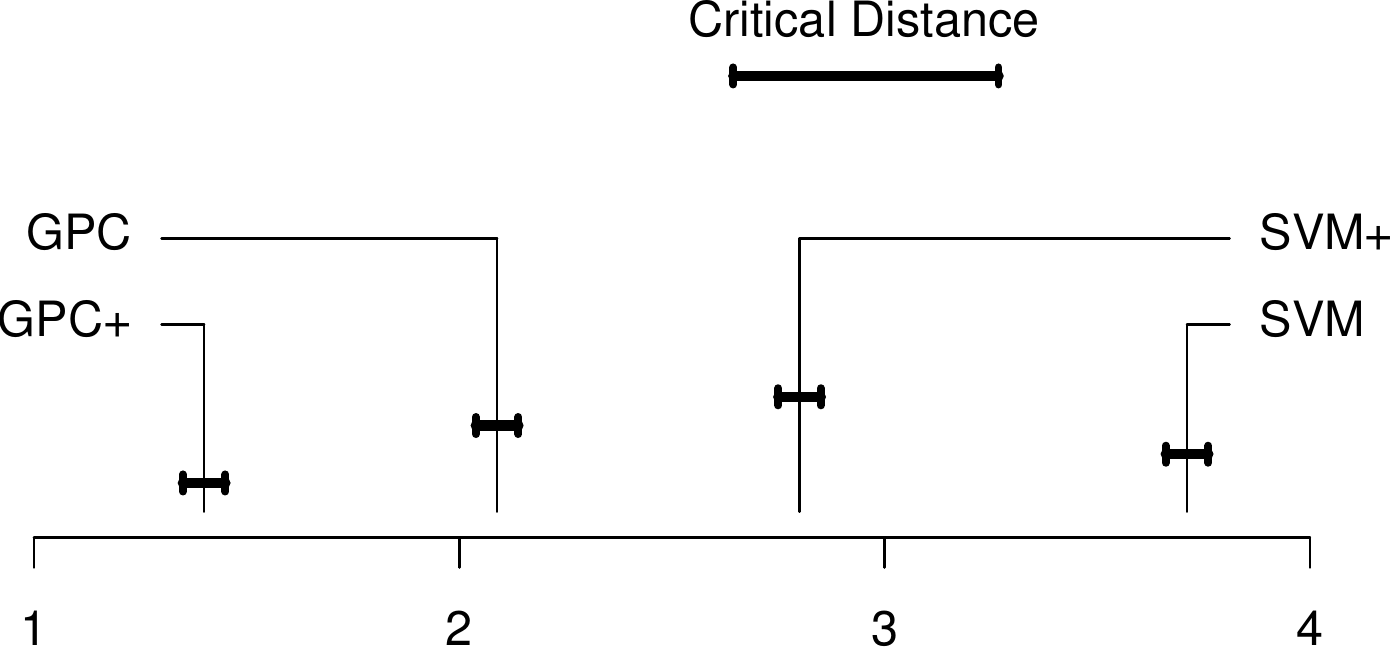} \quad\quad&\quad
    \includegraphics[width=0.45\columnwidth]{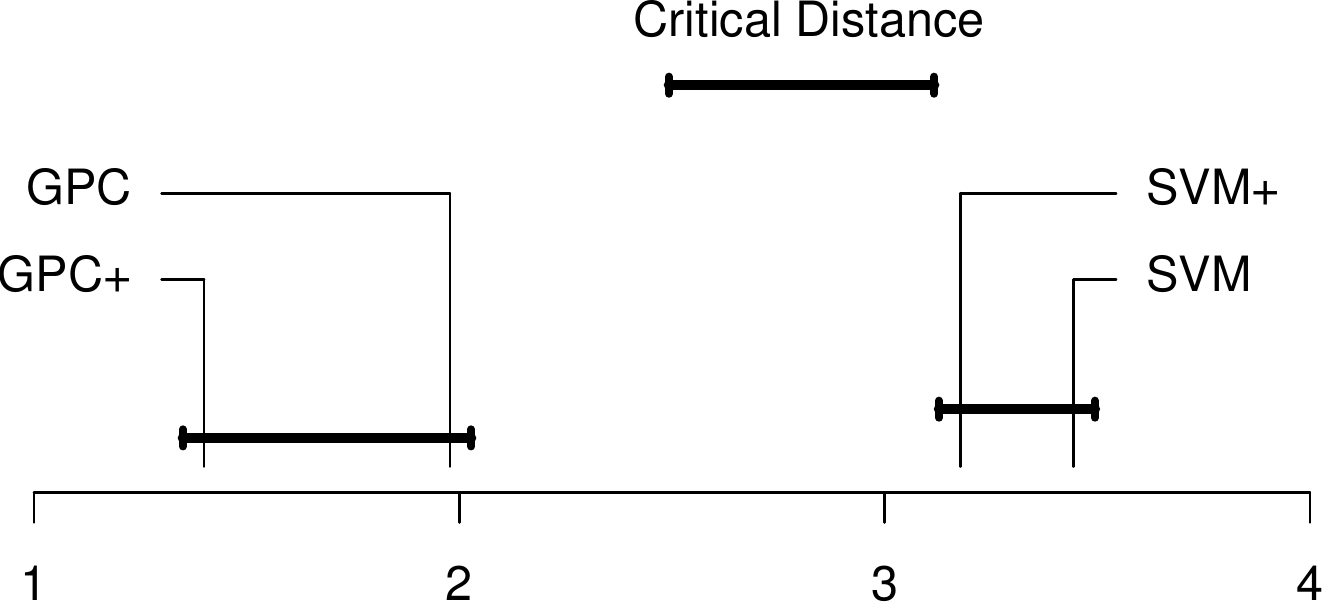}\\
    (DeCAF as privileged) & (Attributes as privileged)
 \end{tabular}
  \caption{Average rank (the lower the better) of the four methods and a critical distance for significant differences \cite{Demsar06} on the AwA dataset. 
  An average rank equal to one means that particular method has the smallest error on the $45$ tasks. 
  Whenever the average ranks differ by more than the critical distance, 
  there is a statistical evidence ($p$-value $< 10\%$) to support a difference in the average ranks and thus in the performance. 
  We also link two methods with a solid line if they are not statistically different from each other ($p$-value $> 10\%$).
  In DeCAF, there is statistical evidence that GPC+ performs best among the four methods considered, 
  while in attributes, GPC+ still performs best but there is not enough evidence to reject that GPC+ performs comparable to GPC. \label{fig:demsar}}
\end{center}
\end{figure*}

{\bf Attributes as privileged information}
Following the experimental setting of \cite{ShaQuaLam13}, we also used \emph{images} as the \emph{original} domain and \emph{attributes} as the \emph{privileged} domain.
Images were represented by $2000$ visual words based on SURF descriptors and 
attributes were in the form of 85 dimensional predicted attributes based on probabilistic binary classifiers \cite{LamNicHar14}.
This time, we only performed PCA and kept the top $50$ principal components in the original domain. 
Finally, we also standardised the data. 

The results are summarised in Fig.~\ref{fig:pairwise} in term of pairwise comparison over $45$ binary tasks between GPC+ and main baselines, GPC and SVM+.   
The full results with the error of each method GPC, GPC+, SVM, and SVM+ on each problem are relegated to 
the appendix.
In contrast to the results on the attribute discovery dataset, on the AwA dataset it is clear that GPC outperforms SVM 
in almost all of the $45$ binary classification tasks (see the appendix). The average error of GPC over $4500$ ($45$ tasks and $100$ repeats per task) experiments is much lower than SVM. On the AwA dataset, SVM+ can take advantage of privileged information -- be it deep belief DeCAF features or semantic attributes -- and shows significant performance improvement over SVM. However, GPC+ still shows the best overall results and further improves the 
already strong performance of GPC. As illustrated in Fig.~\ref{fig:gaussiancdf}~{\bf (right)}, the 
privileged information modulates the slope of the sigmoid likelihood function differently 
for easy and difficult examples: easy examples gain slope and hence importance whereas difficult ones lose importance in the classification. 
We analysed our experimental results using the multiple dataset statistical comparison method described in 
\cite{Demsar06}\footnote{We are not able to use this method for our attribute discovery results in 
Tab.~\ref{tab:ADNNtexts} as the number of methods being compared 
($4$) is almost equal to the number of tasks or datasets ($6$).}. The statistical tests are summarised in 
Fig.~\ref{fig:demsar}.
When DeCAF is used as privileged information, there is statistical evidence that GPC+ \emph{performs best} among the four methods, while in semantic attributes as privileged information setting, GPC+ still performs best but there is not enough evidence to reject that GPC+ performs comparable to GPC.

\section{Conclusions}

We presented the first treatment of the
\emph{learning with privileged information} setting in
the Gaussian process classification (GPC)
framework, called GPC+. 
The privileged information enters the latent noise 
layer of the GPC+, resulting in a data-dependent 
modulation of the sigmoid slope of the GP likelihood. 
As our experimental results demonstrate this 
is an effective way to make use of privileged information, 
which manifest itself in significantly improved classification 
accuracies. Actually, to our knowledge, this is the first time that a 
heteroscedastic noise term is used to improve 
GPC.
Furthermore, we also showed that recent advances in 
continuous word-vector neural networks representations \cite{LeMik14} and deep convolutional networks 
for image representations \cite{DonJiaVinHofetal14} are \emph{privileged} information.
For future work, we plan to extend the GPC+ to the 
multiclass situation and to speed up computation by 
devising a quadrature-free expectation propagation 
method, similar to~\cite{RiiJylVeh13}.

\bibliography{bibfile}
\bibliographystyle{unsrt}

\newpage
\appendix
\section*{Appendix}
Error rate performance on the AwA dataset over 100 repeated experiments. \textbf{SURF} image features as the \textbf{original} domain and \textbf{DeCAF} deep neural network image features as the \textbf{privileged} domain. The best methods for each binary task is highlighted in boldface.
\begin{center}
\scalebox{0.9}{
\begin{tabular}{l*{5}{r@{$\pm$}l}}
      		             & \twocomine{\texttt{GPC}}  & \twocomine{\texttt{GPC+ (Ours)}} & \twocomine{\texttt{SVM}} &  \twocomine{\texttt{SVM+}}\\
\hline\hline
{\tt Chimp. v. Panda} 	& $15.93$ & $0.15$ 	& ${\bf15.86}$ & ${\bf0.15}$ 	& $16.90$ & $0.15$ 	& $16.22$ & $0.14$ 	\\
{\tt Chimp. v. Leopard} & $14.74$ & $0.13$ 	& ${\bf14.69}$ & ${\bf0.12}$ 	& $15.28$ & $0.15$ 	& $15.10$ & $0.12$ 	\\
{\tt Chimp. v. Cat} 	& $16.63$ & $0.12$ 	& ${\bf16.52}$ & ${\bf0.11}$ 	& $17.34$ & $0.11$ 	& $16.97$ & $0.11$ 	\\
{\tt Chimp. v. Pig} 	& $19.82$ & $0.23$ 	& ${\bf19.48}$ & ${\bf0.26}$ 	& $20.52$ & $0.27$ 	& $20.04$ & $0.25$ 	\\
{\tt Chimp. v. Hippo.} 	& $18.99$ & $0.11$ 	& ${\bf18.67}$ & ${\bf0.11}$ 	& $19.57$ & $0.11$ 	& $19.20$ & $0.11$ 	\\
{\tt Chimp. v. Whale} 	& ${\bf5.72}$ & ${\bf0.08}$ 	& $5.79$ & $0.08$ 	& $6.22$ & $0.11$ 	& $5.85$ & $0.10$ 	\\
{\tt Chimp. v. Raccoon} & ${\bf19.60}$ & ${\bf0.13}$ 	& $19.65$ & $0.13$ 	& $20.03$ & $0.14$ 	& $19.86$ & $0.13$ 	\\
{\tt Chimp. v. Rat} 	& $19.94$ & $0.27$ 	& ${\bf19.79}$ & ${\bf0.26}$ 	& $21.53$ & $0.30$ 	& $20.45$ & $0.27$ 	\\
{\tt Chimp. v. Seal} 	& $14.10$ & $0.12$ 	& ${\bf14.01}$ & ${\bf0.12}$ 	& $14.94$ & $0.12$ 	& $14.34$ & $0.13$ 	\\
{\tt Panda v. Leopard} 	& ${\bf14.19}$ & ${\bf0.12}$ 	& $14.26$ & $0.13$ 	& $14.85$ & $0.15$ 	& $14.59$ & $0.15$ 	\\
{\tt Panda v. Cat} 	& ${\bf12.96}$ & ${\bf0.12}$ 	& $12.99$ & $0.12$ 	& $14.26$ & $0.13$ 	& $13.52$ & $0.13$ 	\\
{\tt Panda v. Pig} 	& $20.23$ & $0.23$ 	& ${\bf20.17}$ & ${\bf0.22}$ 	& $22.54$ & $0.29$ 	& $20.85$ & $0.24$ 	\\
{\tt Panda v. Hippo.} 	& $15.47$ & $0.12$ 	& ${\bf15.45}$ & ${\bf0.13}$ 	& $16.89$ & $0.17$ 	& $15.70$ & $0.14$ 	\\
{\tt Panda v. Whale} 	& $5.00$ & $0.09$ 	& $5.20$ & $0.09$ 	& $5.73$ & $0.11$ 	& ${\bf4.95}$ & ${\bf0.09}$ 	\\
{\tt Panda v. Raccoon} 	& ${\bf17.03}$ & ${\bf0.13}$ 	& $17.04$ & $0.13$ 	& $18.37$ & $0.18$ 	& $17.54$ & $0.14$ 	\\
{\tt Panda v. Rat} 	& ${\bf17.09}$ & ${\bf0.24}$ 	& $17.27$ & $0.23$ 	& $20.06$ & $0.30$ 	& $18.04$ & $0.25$ 	\\
{\tt Panda v. Seal} 	& $14.13$ & $0.13$ 	& ${\bf14.01}$ & ${\bf0.12}$ 	& $15.16$ & $0.17$ 	& $14.22$ & $0.13$ 	\\
{\tt Leopard v. Cat} 	& $12.08$ & $0.11$ 	& ${\bf11.77}$ & ${\bf0.09}$ 	& $11.79$ & $0.08$ 	& $11.82$ & $0.09$ 	\\
{\tt Leopard v. Pig} 	& ${\bf20.97}$ & ${\bf0.24}$ 	& $21.12$ & $0.24$ 	& $21.88$ & $0.24$ 	& $21.79$ & $0.26$ 	\\
{\tt Leopard v. Hippo.} & $16.10$ & $0.14$ 	& ${\bf15.89}$ & ${\bf0.13}$ 	& $16.36$ & $0.13$ 	& $16.27$ & $0.14$ 	\\
{\tt Leopard v. Whale} 	& $4.85$ & $0.07$ 	& ${\bf4.79}$ & ${\bf0.07}$ 	& $4.95$ & $0.08$ 	& $4.89$ & $0.07$ 	\\
{\tt Leopard v. Raccoon} & $26.43$ & $0.17$ 	& ${\bf26.24}$ & ${\bf0.16}$ 	& $26.93$ & $0.20$ 	& $26.74$ & $0.19$ 	\\
{\tt Leopard v. Rat} 	& ${\bf17.62}$ & ${\bf0.20}$ 	& $17.75$ & $0.21$ 	& $18.78$ & $0.25$ 	& $18.00$ & $0.23$ 	\\
{\tt Leopard v. Seal} 	& $13.26$ & $0.10$ 	& ${\bf13.05}$ & ${\bf0.10}$ 	& $13.33$ & $0.10$ 	& $13.35$ & $0.10$ 	\\
{\tt Cat v. Pig} 	& $24.35$ & $0.24$ 	& ${\bf23.95}$ & ${\bf0.22}$ 	& $24.52$ & $0.24$ 	& $24.53$ & $0.25$ 	\\
{\tt Cat v. Hippo.} 	& $15.02$ & $0.10$ 	& ${\bf14.85}$ & ${\bf0.10}$ 	& $15.52$ & $0.12$ 	& $15.26$ & $0.11$ 	\\
{\tt Cat v. Whale} 	& $7.66$ & $0.09$ 	& $7.63$ & $0.09$ 	& $7.51$ & $0.09$ 	& ${\bf7.41}$ & ${\bf0.10}$ 	\\
{\tt Cat v. Raccoon} 	& $15.54$ & $0.11$ 	& ${\bf15.35}$ & ${\bf0.10}$ 	& $15.80$ & $0.10$ 	& $15.71$ & $0.13$ 	\\
{\tt Cat v. Rat} 	& $34.96$ & $0.30$ 	& $34.30$ & $0.30$ 	& $34.83$ & $0.32$ 	& ${\bf34.07}$ & ${\bf0.32}$ 	\\
{\tt Cat v. Seal} 	& $18.79$ & $0.14$ 	& ${\bf18.58}$ & ${\bf0.13}$ 	& $18.97$ & $0.14$ 	& $18.85$ & $0.13$ 	\\
{\tt Pig v. Hippo.} 	& $27.57$ & $0.27$ 	& ${\bf27.27}$ & ${\bf0.27}$ 	& $27.75$ & $0.25$ 	& $27.72$ & $0.27$ 	\\
{\tt Pig v. Whale} 	& $8.37$ & $0.15$ 	& ${\bf8.22}$ & ${\bf0.16}$ 	& $8.51$ & $0.17$ 	& $8.24$ & $0.16$ 	\\
{\tt Pig v. Raccoon} 	& $24.45$ & $0.24$ 	& ${\bf24.23}$ & ${\bf0.23}$ 	& $24.40$ & $0.25$ 	& $24.34$ & $0.25$ 	\\
{\tt Pig v. Rat} 	& $30.00$ & $0.26$ 	& ${\bf29.62}$ & ${\bf0.26}$ 	& $30.77$ & $0.30$ 	& $30.22$ & $0.28$ 	\\
{\tt Pig v. Seal} 	& $23.91$ & $0.24$ 	& ${\bf23.37}$ & ${\bf0.21}$ 	& $24.35$ & $0.23$ 	& $23.79$ & $0.22$ 	\\
{\tt Hippo. v. Whale} 	& $14.03$ & $0.12$ 	& ${\bf13.80}$ & ${\bf0.12}$ 	& $14.01$ & $0.12$ 	& $14.01$ & $0.10$ 	\\
{\tt Hippo. v. Raccoon} & $19.31$ & $0.14$ 	& ${\bf19.14}$ & ${\bf0.13}$ 	& $19.61$ & $0.15$ 	& $19.52$ & $0.13$ 	\\
{\tt Hippo. v. Rat} 	& ${\bf21.49}$ & ${\bf0.27}$ 	& $21.82$ & $0.26$ 	& $22.67$ & $0.26$ 	& $22.44$ & $0.27$ 	\\
{\tt Hippo. v. Seal} 	& $30.68$ & $0.17$ 	& ${\bf30.60}$ & ${\bf0.18}$ 	& $31.52$ & $0.18$ 	& $31.03$ & $0.19$ 	\\
{\tt Whale v. Raccoon} 	& $7.92$ & $0.09$ 	& $7.77$ & $0.08$ 	& ${\bf7.61}$ & ${\bf0.09}$ 	& $7.68$ & $0.09$ 	\\
{\tt Whale v. Rat} 	& ${\bf10.98}$ & ${\bf0.22}$ 	& $11.14$ & $0.22$ 	& $11.38$ & $0.24$ 	& $11.17$ & $0.24$ 	\\
{\tt Whale v. Seal} 	& $18.57$ & $0.16$ 	& ${\bf18.18}$ & ${\bf0.16}$ 	& $18.58$ & $0.18$ 	& $18.37$ & $0.16$ 	\\
{\tt Raccoon v. Rat} 	& ${\bf25.16}$ & ${\bf0.27}$ 	& $25.22$ & $0.25$ 	& $25.90$ & $0.24$ 	& $25.73$ & $0.25$ 	\\
{\tt Raccoon v. Seal} 	& $15.06$ & $0.13$ 	& ${\bf14.82}$ & ${\bf0.13}$ 	& $15.43$ & $0.12$ 	& $15.35$ & $0.12$ 	\\
{\tt Rat v. Seal} 	& $24.91$ & $0.28$ 	& ${\bf24.63}$ & ${\bf0.27}$ 	& $25.24$ & $0.28$ 	& $25.16$ & $0.28$ 	\\	
\hline\hline
{\it average ranking}            & \twocomine{2.09} & \twocomine{1.40} &    \twocomine{ 3.71} & \twocomine{2.80}\\
{\it average error}      & 17.60&0.10 & {\bf 17.47}&{\bf 0.10} &  18.21&0.11 & 17.80 &0.10
\label{tab:AwA}
\end{tabular}}
\end{center}
\noindent
Error rate performance on the AwA dataset over 100 repeated experiments. \textbf{SURF} image features as the \textbf{original} domain and \textbf{attributes} as the \textbf{privileged} domain. The best methods for each binary task is highlighted in boldface
\begin{center}
\scalebox{0.9}{
\begin{tabular}{l*{5}{r@{$\pm$}l}}
      		             & \twocomine{\texttt{GPC}}  & \twocomine{\texttt{GPC+ (Ours)}} & \twocomine{\texttt{SVM}} &  \twocomine{\texttt{SVM+}}\\
\hline\hline
{\tt Chimp. v. Panda} 	& $15.93$ & $0.15$ 	& ${\bf15.85}$ & ${\bf0.14}$ 	& $16.90$ & $0.15$ 	& $16.64$ & $0.15$ 	\\
{\tt Chimp. v. Leopard} & $14.74$ & $0.13$ 	& ${\bf14.72}$ & ${\bf0.12}$ 	& $15.28$ & $0.15$ 	& $15.18$ & $0.12$ 	\\
{\tt Chimp. v. Cat} 	& $16.63$ & $0.12$ 	& ${\bf16.51}$ & ${\bf0.11}$ 	& $17.34$ & $0.11$ 	& $17.08$ & $0.10$ 	\\
{\tt Chimp. v. Pig} 	& $19.82$ & $0.23$ 	& ${\bf19.50}$ & ${\bf0.26}$ 	& $20.52$ & $0.27$ 	& $20.34$ & $0.24$ 	\\
{\tt Chimp. v. Hippo.} 	& $18.99$ & $0.11$ 	& ${\bf18.68}$ & ${\bf0.11}$ 	& $19.57$ & $0.11$ 	& $19.36$ & $0.11$ 	\\
{\tt Chimp. v. Whale} 	& $5.72$ & $0.08$ 	& $5.79$ & $0.08$ 	& $6.22$ & $0.11$ 	& ${\bf5.63}$ & ${\bf0.08}$ 	\\
{\tt Chimp. v. Raccoon} & ${\bf19.60}$ & ${\bf0.13}$ 	& $19.66$ & $0.13$ 	& $20.03$ & $0.14$ 	& $20.13$ & $0.13$ 	\\
{\tt Chimp. v. Rat} 	& $19.94$ & $0.27$ 	& ${\bf19.82}$ & ${\bf0.26}$ 	& $21.53$ & $0.30$ 	& $20.22$ & $0.28$ 	\\
{\tt Chimp. v. Seal} 	& $14.10$ & $0.12$ 	& ${\bf14.03}$ & ${\bf0.12}$ 	& $14.94$ & $0.12$ 	& $14.57$ & $0.15$ 	\\
{\tt Panda v. Leopard} 	& ${\bf14.19}$ & ${\bf0.12}$ 	& $14.21$ & $0.13$ 	& $14.85$ & $0.15$ 	& $14.82$ & $0.14$ 	\\
{\tt Panda v. Cat} 	& ${\bf12.96}$ & ${\bf0.12}$ 	& $12.98$ & $0.12$ 	& $14.26$ & $0.13$ 	& $13.96$ & $0.12$ 	\\
{\tt Panda v. Pig} 	& $20.23$ & $0.23$ 	& ${\bf20.18}$ & ${\bf0.22}$ 	& $22.54$ & $0.29$ 	& $21.41$ & $0.26$ 	\\
{\tt Panda v. Hippo.} 	& $15.47$ & $0.12$ 	& ${\bf15.44}$ & ${\bf0.13}$ 	& $16.89$ & $0.17$ 	& $16.16$ & $0.17$ 	\\
{\tt Panda v. Whale} 	& ${\bf5.00}$ & ${\bf0.09}$ 	& $5.20$ & $0.09$ 	& $5.73$ & $0.11$ 	& $5.06$ & $0.08$ 	\\
{\tt Panda v. Raccoon} 	& $17.03$ & $0.13$ 	& ${\bf17.02}$ & ${\bf0.13}$ 	& $18.37$ & $0.18$ 	& $18.06$ & $0.15$ 	\\
{\tt Panda v. Rat} 	& ${\bf17.09}$ & ${\bf0.24}$ 	& $17.26$ & $0.23$ 	& $20.06$ & $0.30$ 	& $18.37$ & $0.25$ 	\\
{\tt Panda v. Seal} 	& $14.13$ & $0.13$ 	& ${\bf13.99}$ & ${\bf0.13}$ 	& $15.16$ & $0.17$ 	& $14.82$ & $0.15$ 	\\
{\tt Leopard v. Cat} 	& $12.08$ & $0.11$ 	& $11.81$ & $0.09$ 	& ${\bf11.79}$ & ${\bf0.08}$ 	& $12.13$ & $0.11$ 	\\
{\tt Leopard v. Pig} 	& ${\bf20.97}$ & ${\bf0.24}$ 	& $21.11$ & $0.24$ 	& $21.88$ & $0.24$ 	& $22.03$ & $0.29$ 	\\
{\tt Leopard v. Hippo.} & $16.10$ & $0.14$ 	& ${\bf15.87}$ & ${\bf0.13}$ 	& $16.36$ & $0.13$ 	& $16.41$ & $0.14$ 	\\
{\tt Leopard v. Whale} 	& $4.85$ & $0.07$ 	& ${\bf4.78}$ & ${\bf0.07}$ 	& $4.95$ & $0.08$ 	& $4.90$ & $0.07$ 	\\
{\tt Leopard v. Raccoon} & $26.43$ & $0.17$ 	& ${\bf26.27}$ & ${\bf0.17}$ 	& $26.93$ & $0.20$ 	& $27.31$ & $0.19$ 	\\
{\tt Leopard v. Rat} 	& ${\bf17.62}$ & ${\bf0.20}$ 	& $17.67$ & $0.21$ 	& $18.78$ & $0.25$ 	& $18.84$ & $0.24$ 	\\
{\tt Leopard v. Seal} 	& $13.26$ & $0.10$ 	& ${\bf13.04}$ & ${\bf0.10}$ 	& $13.33$ & $0.10$ 	& $13.38$ & $0.10$ 	\\
{\tt Cat v. Pig} 	& $24.35$ & $0.24$ 	& ${\bf24.01}$ & ${\bf0.23}$ 	& $24.52$ & $0.24$ 	& $24.68$ & $0.25$ 	\\
{\tt Cat v. Hippo.} 	& $15.02$ & $0.10$ 	& ${\bf14.87}$ & ${\bf0.10}$ 	& $15.52$ & $0.12$ 	& $15.47$ & $0.11$ 	\\
{\tt Cat v. Whale} 	& $7.66$ & $0.09$ 	& $7.63$ & $0.09$ 	& $7.51$ & $0.09$ 	& ${\bf7.48}$ & ${\bf0.09}$ 	\\
{\tt Cat v. Raccoon} 	& $15.54$ & $0.11$ 	& ${\bf15.32}$ & ${\bf0.10}$ 	& $15.80$ & $0.10$ 	& $15.78$ & $0.10$ 	\\
{\tt Cat v. Rat} 	& $34.96$ & $0.30$ 	& ${\bf34.42}$ & ${\bf0.29}$ 	& $34.83$ & $0.32$ 	& $34.79$ & $0.28$ 	\\
{\tt Cat v. Seal} 	& $18.79$ & $0.14$ 	& ${\bf18.60}$ & ${\bf0.13}$ 	& $18.97$ & $0.14$ 	& $19.25$ & $0.16$ 	\\
{\tt Pig v. Hippo.} 	& $27.57$ & $0.27$ 	& ${\bf27.36}$ & ${\bf0.26}$ 	& $27.75$ & $0.25$ 	& $28.17$ & $0.26$ 	\\
{\tt Pig v. Whale} 	& $8.37$ & $0.15$ 	& ${\bf8.24}$ & ${\bf0.16}$ 	& $8.51$ & $0.17$ 	& $8.38$ & $0.15$ 	\\
{\tt Pig v. Raccoon} 	& $24.45$ & $0.24$ 	& ${\bf24.24}$ & ${\bf0.22}$ 	& $24.40$ & $0.25$ 	& $24.91$ & $0.22$ 	\\
{\tt Pig v. Rat} 	& $30.00$ & $0.26$ 	& ${\bf29.67}$ & ${\bf0.26}$ 	& $30.77$ & $0.30$ 	& $30.33$ & $0.29$ 	\\
{\tt Pig v. Seal} 	& $23.91$ & $0.24$ 	& ${\bf23.41}$ & ${\bf0.22}$ 	& $24.35$ & $0.23$ 	& $24.03$ & $0.23$ 	\\
{\tt Hippo. v. Whale} 	& $14.03$ & $0.12$ 	& ${\bf13.82}$ & ${\bf0.12}$ 	& $14.01$ & $0.12$ 	& $14.45$ & $0.13$ 	\\
{\tt Hippo. v. Raccoon} & $19.31$ & $0.14$ 	& ${\bf19.15}$ & ${\bf0.13}$ 	& $19.61$ & $0.15$ 	& $19.74$ & $0.16$ 	\\
{\tt Hippo. v. Rat} 	& ${\bf21.49}$ & ${\bf0.27}$ 	& $21.79$ & $0.26$ 	& $22.67$ & $0.26$ 	& $22.33$ & $0.26$ 	\\
{\tt Hippo. v. Seal} 	& $30.68$ & $0.17$ 	& ${\bf30.61}$ & ${\bf0.18}$ 	& $31.52$ & $0.18$ 	& $31.47$ & $0.19$ 	\\
{\tt Whale v. Raccoon} 	& $7.92$ & $0.09$ 	& $7.79$ & $0.08$ 	& ${\bf7.61}$ & ${\bf0.09}$ 	& $7.66$ & $0.08$ 	\\
{\tt Whale v. Rat} 	& ${\bf10.98}$ & ${\bf0.22}$ 	& $11.16$ & $0.22$ 	& $11.38$ & $0.24$ 	& $11.17$ & $0.22$ 	\\
{\tt Whale v. Seal} 	& $18.57$ & $0.16$ 	& ${\bf18.28}$ & ${\bf0.16}$ 	& $18.58$ & $0.18$ 	& $18.99$ & $0.18$ 	\\
{\tt Raccoon v. Rat} 	& ${\bf25.16}$ & ${\bf0.27}$ 	& $25.21$ & $0.25$ 	& $25.90$ & $0.24$ 	& $26.12$ & $0.26$ 	\\
{\tt Raccoon v. Seal} 	& $15.06$ & $0.13$ 	& ${\bf14.85}$ & ${\bf0.12}$ 	& $15.43$ & $0.12$ 	& $15.42$ & $0.13$ 	\\
{\tt Rat v. Seal} 	& $24.91$ & $0.28$ 	& ${\bf24.64}$ & ${\bf0.26}$ 	& $25.24$ & $0.28$ 	& $25.14$ & $0.27$ 	\\
\hline\hline
{\it average ranking}            & \twocomine{1.98} &  \twocomine{1.40 }    &    \twocomine{3.44 }&  \twocomine{3.18 }\\
{\it average error}      & 17.60&0.10 & {\bf17.48}&{\bf0.10} & 18.21&0.11 & 18.06&0.11
\label{tab:AwA}
\end{tabular}}
\end{center}

\end{document}

%% file: definitions.tex
\usepackage{amssymb}
\usepackage{latexsym}
\usepackage{amsfonts}
\usepackage{amsmath}

\usepackage{theorem}
\newcommand{\BlackBox}{ule{1.5ex}{1.5ex}}  

\usepackage[tight,normalsize,sf,SF]{subfigure}

\usepackage{algorithm}
\usepackage{algorithmic}

\usepackage{xcolor}
\usepackage{stmaryrd}

\newcommand{\eq}[1]{(\ref{#1})}

\newcommand{\nbr}[1]{\left\|#1\right\|}

\newcommand{\RR}{\mathbb{R}}

\newcommand{\Ncal}{\mathcal{N}}

\newcommand{\Dcal}{\mathcal{D}}

\newcommand{\x}{\mathbf{x}}
\newcommand{\y}{\mathbf{y}}

\newcommand{\pr}{\text{Pr}}

\newcommand{\I}{\mathbb{I}}

\newcommand{\tildef}{\tilde{f}}

\newcommand{\bftildef}{\mathbf{\tilde{f}}}

\usepackage[normalem]{ulem}

\usepackage{xspace} 
\makeatletter
\DeclareRobustCommand\onedot{\futurelet\@let@token\@onedot}
\def\@onedot{\ifx\@let@token.\else.\null\fi\xspace}

\def\eg{\emph{e.g}\onedot}

\def\eg{{e.g}\onedot}